\documentclass[10pt,twocolumn,letterpaper]{article}

 \usepackage{cvpr}              

\usepackage{tablefootnote}
\usepackage{multirow}
\usepackage{booktabs}
\usepackage{multirow}
\usepackage{amsfonts}
\usepackage[table,xcdraw]{xcolor}
\usepackage{amsmath}
\usepackage{amssymb}
\usepackage{siunitx}
\usepackage{graphicx}
\usepackage{subcaption}
\usepackage[accsupp]{axessibility}

\definecolor{cvprblue}{rgb}{0.21,0.49,0.74}
\usepackage[pagebackref,breaklinks,colorlinks,citecolor=cvprblue]{hyperref}

\def\paperID{5807} 
\def\confName{CVPR}
\def\confYear{2024}

\title{DuPL: Dual Student with Trustworthy Progressive Learning for Robust \\ Weakly Supervised Semantic Segmentation}

\author{Yuanchen Wu$^{1}$, Xichen Ye$^{1}$, Kequan Yang$^{1}$, Jide Li$^{1}$, Xiaoqiang Li$^{1}$\thanks{Corresponding author.}
\\
$^{1}$ School of Computer Engineering and Science, Shanghai University, China.\\
{\tt\small \{yuanchenwu,yexichen0930,kqyang,iavtvai,xqli\}@shu.edu.cn}
}

\begin{document}
\maketitle

\begin{abstract}
Recently, One-stage Weakly Supervised Semantic Segmentation (WSSS) with image-level labels has gained increasing interest due to simplification over its cumbersome multi-stage counterpart. Limited by the inherent ambiguity of Class Activation Map (CAM), we observe that one-stage pipelines often encounter confirmation bias caused by incorrect CAM pseudo-labels, impairing their final segmentation performance. Although recent works discard many unreliable pseudo-labels to implicitly alleviate this issue, they fail to exploit sufficient supervision for their models. To this end, we propose a dual student framework with trustworthy progressive learning (\textbf{DuPL}). Specifically, we propose a dual student network with a discrepancy loss to yield diverse CAMs for each sub-net. The two sub-nets generate supervision for each other, mitigating the confirmation bias caused by learning their own incorrect pseudo-labels. In this process, we progressively introduce more trustworthy pseudo-labels to be involved in the supervision through dynamic threshold adjustment with an adaptive noise filtering strategy. Moreover, we believe that every pixel, even discarded from supervision due to its unreliability, is important for WSSS. Thus, we develop consistency regularization on these discarded regions, providing supervision of every pixel. Experiment results demonstrate the superiority of the proposed DuPL over the recent state-of-the-art alternatives on PASCAL VOC 2012 and MS COCO datasets. Code is available at \url{https://github.com/Wu0409/DuPL} .

\end{abstract}

\begin{figure}[!tp]
    \centering
    \includegraphics[width=0.43\textwidth]{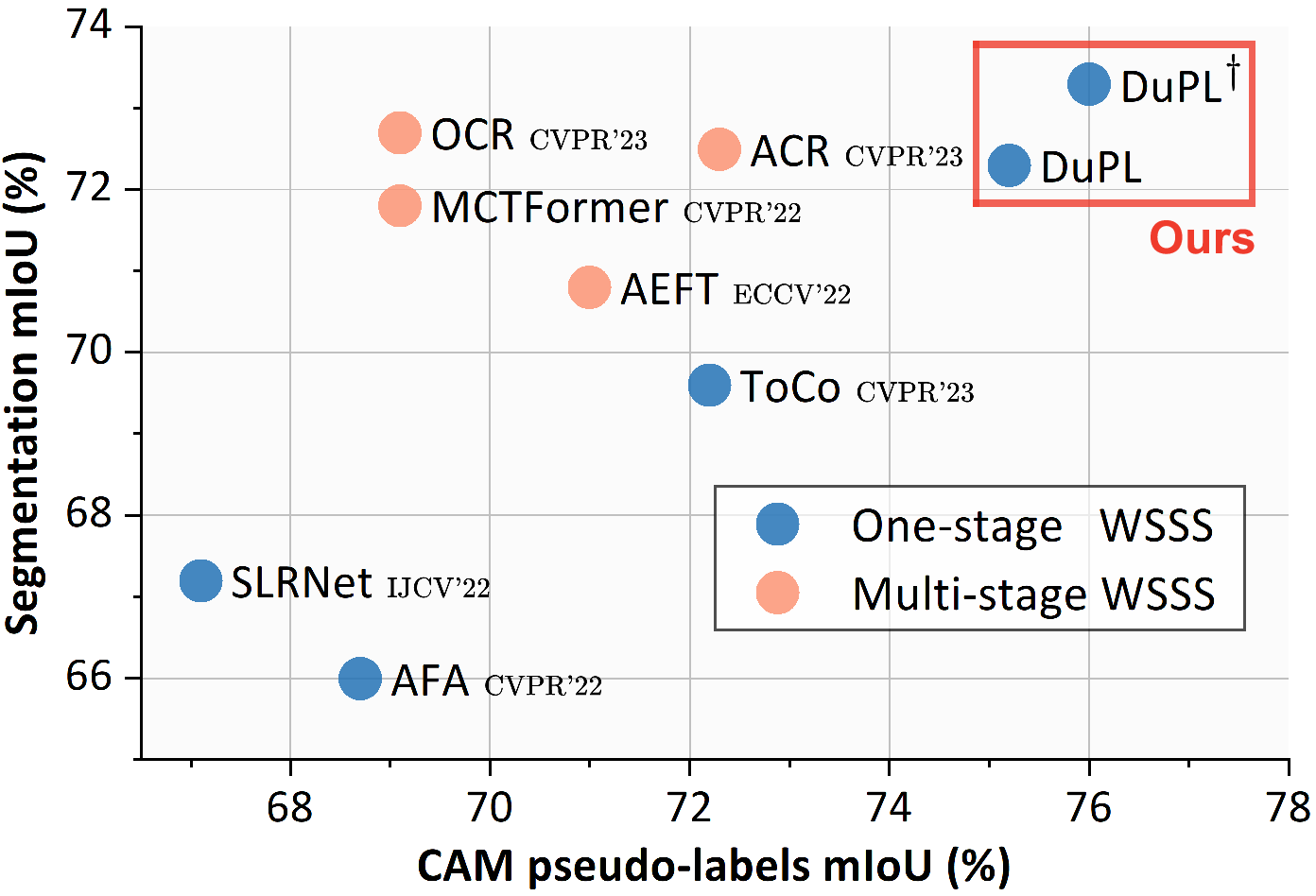}
    \vspace{-2mm}
    \caption{\textbf{CAM pseudo-labels (\textit{train}) \textit{vs.} segmentation performance (\textit{val}) on PASCAL VOC 2012}. DuPL outperforms state-of-the-art one-stage competitors and achieves comparable performance with multi-stage methods in terms of CAM pseudo-labels and final segmentation performance. $\dagger$ denotes using ImageNet-21k pretrained parameters.}
    \vspace{-6mm}
    \label{fig_intro}
\end{figure}

\section{Introduction}
Weakly supervised semantic segmentation (WSSS) aims at using weak supervision, such as image-level labels \cite{du2022weakly, ijcai2023p171}, scribbles \cite{lin2016scribblesup, vernaza2017learning}, and bounding boxes \cite{lee2021bbam, oh2021background}, to alleviate the reliance on pixel-level annotations for segmentation. Among these annotation forms, using image-level labels is the most rewarding yet challenging way, as it only provides the presence of certain classes without offering any localization information. In this paper, we also focus on semantic segmentation using image-level labels.
   
Prevalent works typically follow a multi-stage pipeline \cite{kolesnikov2016seed}, \ie, pseudo-label generation, refinement, and segmentation training. First, the pixel-level pseudo-labels are derived from Class Activation Map (CAM) through classification \cite{zhou2016learning}. Since CAM tends to identify the discriminative semantic regions and fails to distinguish co-occurring objects, the pseudo-labels often suffer from the CAM ambiguity. Thus, they are then refined by training a refinement network \cite{ahn2018learning, ahn2019weakly}. Finally, the refined pseudo-labels are used to train a segmentation model in a fully supervised manner. Recently, to simplify the multi-stage process, many studies proposed one-stage solutions that simultaneously produce pseudo-labels and learn a segmentation head \cite{araslanov2020single, ru2022learning, ru2023token}. Despite their enhanced training efficiency, the performance still lags behind their multi-stage counterparts.

\begin{figure}[!tp]
    \centering
    \includegraphics[width=0.43\textwidth]{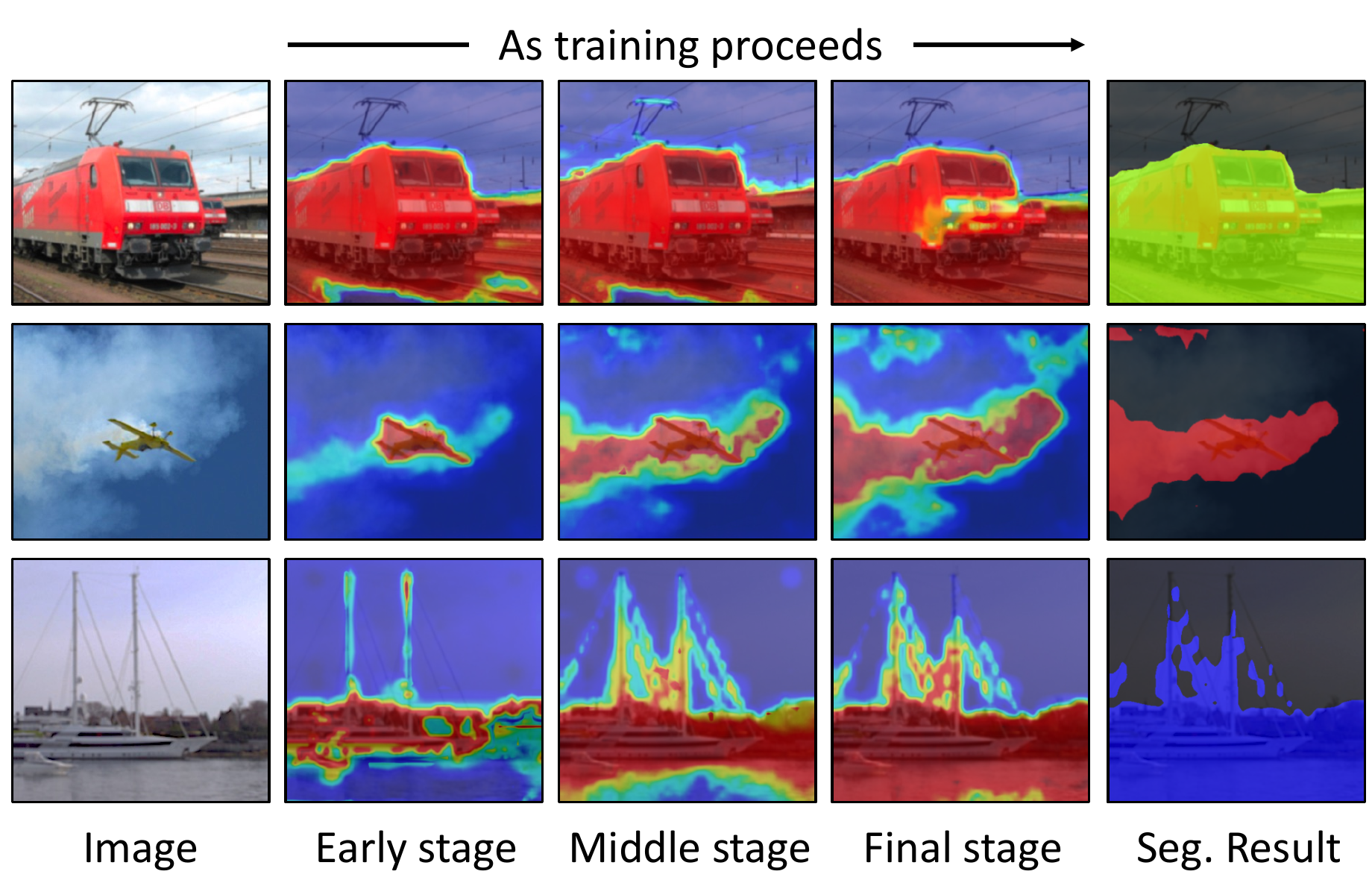}
    \vspace{-2mm}
    \caption{\textbf{Confirmation bias of CAM.} As training proceeds, the bias will be consistently reinforced, impairing the final segmentation performance. Here, we use the ViT-B \cite{dosovitskiy2020image} baseline and introduce more unreliable pseudo-labels to amplify this phenomenon.}
    \vspace{-6mm}
    \label{fig_bias}
\end{figure}

One important yet overlooked reason is the confirmation bias of CAM, stemming from the concurrent process of CAM pseudo-label generation and segmentation supervision. For the one-stage pipeline, the segmentation training enforces the backbone features to align with the CAM pseudo-labels. Since the backbone features are shared for the segmentation head and the CAM generation, these inaccurate CAM pseudo-labels not only hinder the learning process of segmentation but, more critically, reinforce the CAM's incorrect judgments. As illustrated in Figure \ref{fig_bias}, this issue consistently deteriorates throughout the training phase and eventually degrades the segmentation performance. Recent one-stage approaches \cite{ru2022learning, xu2023self, ru2023token} commonly set a fixed and high threshold to filter unreliable pseudo-labels, which prioritizes high-quality supervision to implicitly alleviate this issue. However, this strategy fails to exploit sufficient supervision for their models. Employing a fixed and high threshold inevitably discards many pixels that actually have correct CAM pseudo-labels. Furthermore, these unreliable regions discarded from supervision often exist in semantically ambiguous regions. Excluding them directly from supervision makes the model rarely learn the segmentation in these regions, leading to insufficient training. From this perspective, we believe that every pixel matters for segmentation and should be properly utilized.

To address the above limitations, this work proposes a \textit{dual student framework with trustworthy progressive learning}, dubbed DuPL. Inspired by the co-training \cite{qiao2018deep} paradigm, we equip two student sub-networks that engage in mutual learning. They infer diverse CAMs from different views, and transfer the knowledge learned from one view to the other. To avoid homogenous students, we impose a representation-level discrepancy constraint on the two sub-nets. This architecture effectively mitigates the confirmation bias resulting from their own incorrect pseudo-labels, thus producing high-fidelity CAMs. Based on our dual student framework, we propose trustworthy progressive learning for sufficient segmentation supervision. We set up a dynamic threshold adjustment strategy to involve more pixels in the segmentation supervision. To overcome the noise in CAM pseudo-labels, we propose an adaptive noise filtering strategy based on the Gaussian Mixture Model. Finally, for the regions where pseudo-labels are excluded from supervision due to their unreliability, we employ an additional strong perturbation branch for each sub-net and develop consistency regularization on these regions. Overall, our main contributions are summarized as follows:

\begin{itemize}
	\item We explore the CAM confirmation bias in one-stage WSSS. To address this limitation, we propose a dual student architecture. Our experiment proves its effectiveness of reducing the over-activation rate caused by this issue and promotes the quality of CAM pseudo-labels.
	\vspace{2pt}
	\item We propose progressive learning with adaptive noise filtering, which allows more trustworthy pixels to participate in supervision. For the regions with filtered pseudo-labels, we develop consistency regularization for sufficient training. This strategy highlights the importance of fully exploiting pseudo-supervision for WSSS.
	\vspace{2pt}
	\item Experiments on the PASCAL VOC and MS COCO datasets show that DuPL surpasses state-of-the-art one-stage WSSS competitors and achieves comparable performance with multi-stage solutions (Figure \ref{fig_intro}). Through visualizing the segmentation results, we observe that DuPL shows much better segmentation robustness, thanks to our dual student and trust-worthy progressive learning.
\end{itemize}

\section{Related work}
\noindent\textbf{One-stage Weakly Supervised Semantic Segmentation.} Due to the complex process of multi-stage solutions \cite{ahn2018learning, ahn2019weakly}, many recent efforts mainly focused on one-stage solutions \cite{araslanov2020single, ru2022learning,xu2023self,ru2023token}. A common one-stage pipeline is generating CAM and using online refinement modules to obtain final pseudo-labels \cite{araslanov2020single}. These pseudo-labels are then directly used as the supervision for the segmentation head. Typically, recent works mainly proposed additional modules or training objectives to achieve better segmentation. For instance, Zhang \etal \cite{zhang2020reliability} introduce a feature-to-prototype alignment loss with an adaptive affinity field, Ru \etal \cite{ru2022learning} leverage pseudo-labels to guide the affinity learning of self-attention, and Xu \etal \cite{xu2023self} utilize feature correspondence to achieve self-distillation. One common practice of them is that they all set a high and fixed threshold to filter out unreliable pseudo-labels to ensure the quality of supervision. In contrast, we propose a progressive learning strategy fully exploit the potential of every pseudo-label. 

\vspace{2pt}

\noindent\textbf{Confirmation Bias.} This phenomenon commonly occurs in the self-training paradigm of semi-supervised learning (SSL) \cite{lee2013pseudo}, where the model overfits the unlabeled images assigned with incorrect pseudo-labels. In the above process, this incorrect information is constantly reinforced, causing the unstable training process \cite{arazo2020pseudo}. Co-training offers an effective solution to this issue \cite{qiao2018deep}. It uses two diverse sub-nets to provide mutual supervision to ensure more stable and accurate predictions while mitigating the confirmation bias \cite{ouali2020semi,chen2021semi}. Motivated by this, we propose a dual student architecture with a representation-level discrepancy loss to generate diverse CAMs. The two sub-nets learn from each other through the other's pseudo-labels, countering the CAM confirmation bias and achieving better object activation. To the best of our knowledge, DuPL is the first work exploring the CAM confirmation bias in one-stage WSSS.

\vspace{2pt}

\noindent\textbf{Noise Label Learning in WSSS.} In addition to better CAM pseudo-label generation, several recent works aim at learning a robust segmentation model using existing pseudo-labels \cite{li2022uncertainty,liu2022adaptive, cheng2023out}. URN \cite{li2022uncertainty} introduces the uncertainty estimation by the pixel-wise variance between different views to filter noisy labels. Based on the early learning and memorization phenomenon \cite{liu2020early}, ADELE \cite{liu2022adaptive} adaptively calibrates noise labels based on prior outputs in the early learning stage. Different from these works relying on the existing CAM pseudo-labels by other works, the pseudo-labels in one-stage methods continuously update in training. To alleviate the noise pseudo-labels for our progressive learning, we design an online adaptive noise filtering strategy based on the loss feedback from the segmentation head. 

\begin{figure*}[!tp]
    \centering
    \includegraphics[width=0.85\textwidth]{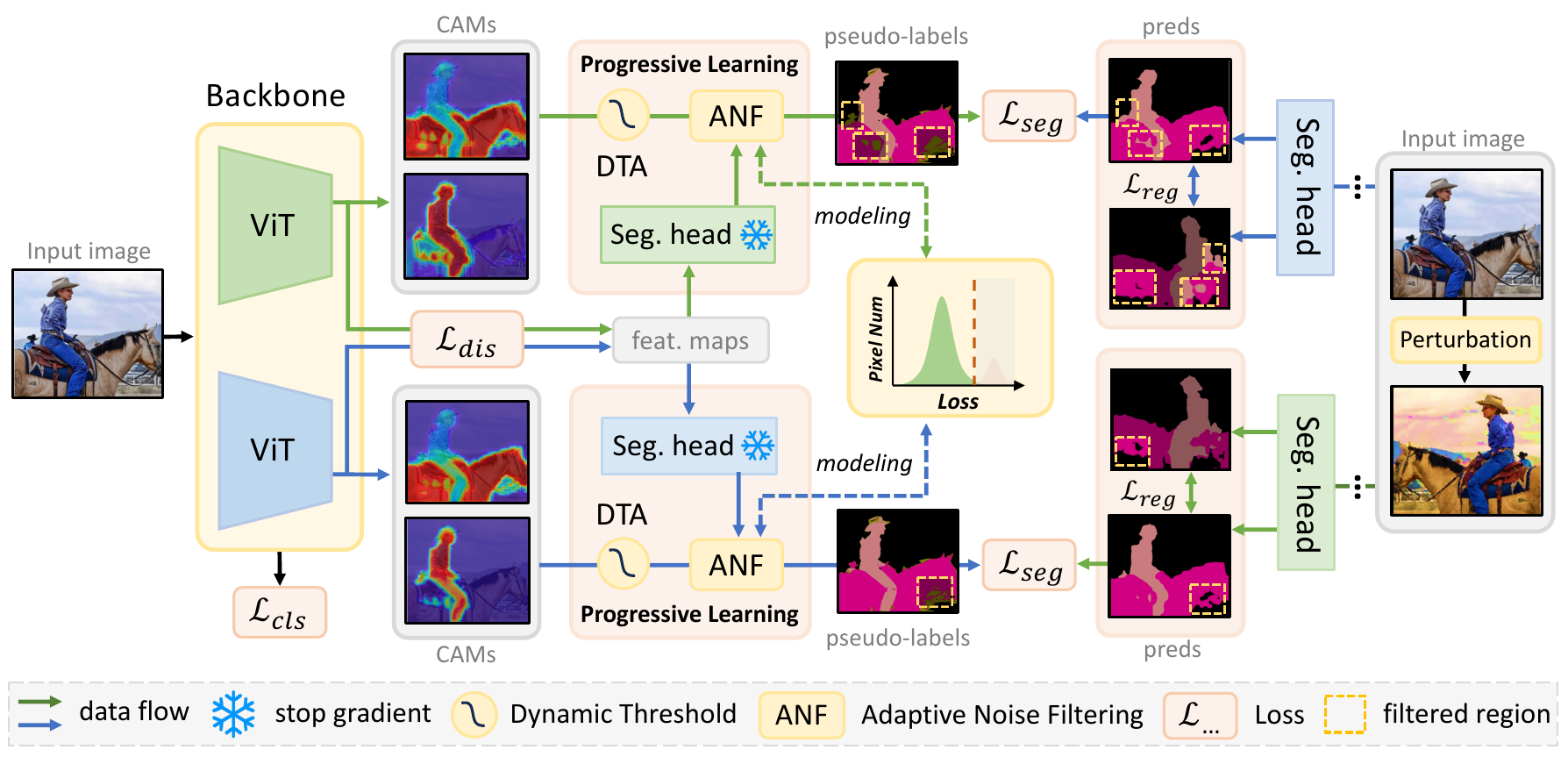}
    \vspace{-2mm}
    \caption{\textbf{The overall framework of DuPL.} We use a discrepancy loss $\mathcal{L}_{dis}$ to constrain the two sub-nets to generate diverse CAMs. Their CAM pseudo-labels are utilized for segmentation cross-supervision $\mathcal{L}_{seg}$, which mitigates the CAM confirmation bias. In this process, we set a dynamic threshold to progressively introduce more pixels to segmentation supervision. Adaptive Noise Filtering strategy is equipped to minimize the noise in pseudo-labels via the segmentation loss distribution. To utilize every pixel, the filtered regions are implemented consistency regularization $\mathcal{L}_{reg}$ with their perturbed counterparts. The classifier is simplified for the clear illustration.} 
    \vspace{-4mm}
    \label{fig_overview}
\end{figure*}

\section{Method}
\subsection{Preliminary}
\label{sect_pre}
We begin with a brief review of how to generate CAM \cite{zhou2016learning} and its pseudo-labels. Given an image, its feature maps $\mathbf{F} \in \mathbb{R}^{D \times H \times W} $are extracted by a backbone network, where $D$ and $H \times W$ is the channel and spatial dimension, respectively. Then, $\mathbf{F}$ is fed to a global average pooling and a classification layer to output the final classification score. In the above process, we can retrieve the classification weight of every class $\mathbf{W} \in \mathbb{R}^{C \times D}$ and use it to weight and sum the feature maps to generate the CAM:
\begin{equation}
\mathbf{M}(c)=\texttt{ReLU}\left(\sum_{i=1}^D \mathbf{W}_{c, i} \cdot \mathbf{F}_i\right),
\end{equation}
where $c$ is the $c$-th class and \texttt{ReLU} is used to eliminate negative activations. Finally, we apply \texttt{max-min} normalization to rescale $\mathbf{M} \in \mathbb{R}^{C \times H \times W}$ to $[0, 1]$. To generate the CAM pseudo-labels, one-stage WSSS methods commonly use two background thresholds $\tau_l$ and $\tau_h$ to separate the background ($\mathbf{M} \leq \tau_l$), uncertain region ($\tau_l < \mathbf{M} < \tau_h$), and foreground ($\mathbf{M} \geq \tau_h$) \cite{ru2022learning, ru2023token}. The uncertain part is regarded as unreliable regions with noise, and will not be involved in the supervision of the segmentation head.

\subsection{Dual Student Framework}
To overcome the confirmation bias of CAM, we propose a co-training-based dual student network where two sub-nets (\ie, $\psi_1$ and $\psi_2$) have the same network architecture, and their parameters are independently updated and non-shared. As presented in Figure \ref{fig_overview}, for the $i$-th sub-net, it comprises a backbone network $\psi^f_i$, a classifier $\psi^{c}_i$, and a segmentation head $\psi^s_i$. To ensure that the two sub-nets activate more diverse regions in CAMs, we enforce sufficient diversity to their representations extracted from $\psi^f_i$, preventing two sub-nets from being homogeneous such that one subnet can learn the knowledge from the other to alleviate the confirmation bias of CAM. Therefore, we set a discrepancy constraint to minimize the cosine similarity between the feature maps of two sub-nets. Formally, denoting the input image as $\mathbf{X}$ and the features from the sub-nets as $\boldsymbol{f}_1=\psi^f_1(\mathbf{X})$ and $\boldsymbol{f}_2=\psi^f_2(\mathbf{X})$, we minimize their similarity by:
\begin{equation}
\mathcal{D}\left(\boldsymbol{f}_1, \boldsymbol{f}_2\right)= -\,log\left(1 - \frac{\boldsymbol{f}_1 \cdot \boldsymbol{f}_2}{\left\|\boldsymbol{f}_1\right\|_2 \times\left\|\boldsymbol{f}_2\right\|_2}\right),
\end{equation}
where $\left\|\cdot\right\|_2$ is $l$2-normalization. Following \cite{grill2020bootstrap,chen2021exploring}, we define a symmetrized discrepancy loss as:
\begin{equation}
\mathcal{L}_{dis}=\mathcal{D}(\boldsymbol{f}_1, \Delta(\boldsymbol{f}_2))+ \mathcal{D}(\boldsymbol{f}_2, \Delta(\boldsymbol{f}_1)),
\end{equation}
where $\Delta$ is the stop-gradient operation to avoid the model from collapse. This loss is computed for each image, with the total loss being the average across all images.

\vspace{1pt}

The segmentation supervision of dual student is bidirectional. One is from $\mathbf{M}_1$ to $\psi_2$  and the other one is $\mathbf{M}_2$ to $\psi_1$, where $\mathbf{M}_1$, $\mathbf{M}_2$ are the CAM from the sub-nets $\psi_1$, $\psi_2$, respectively. The CAM pseudo-labels $\mathbf{Y}_1$ from $\mathbf{M}_1$ are used to supervise the prediction maps $\mathbf{P}_2$ from the other sub-net’s segmentation head $\psi^s_2$, and vice versa. The segmentation loss of our framework is computed as:
\begin{equation}
\mathcal{L}_{seg}=\texttt{CE}(\mathbf{P}_1,\mathbf{Y}_2)+\texttt{CE}(\mathbf{P}_2,\mathbf{Y}_1),
\end{equation}
where \texttt{CE} is the standard cross-entropy loss function.

\subsection{Trustworthy Progressive Learning}
\noindent\textbf{Dynamic Threshold Adjustment.} As mentioned in Section \ref{sect_pre}, one-stage methods \cite{ru2022learning,xu2023self,ru2023token} set background thresholds, $\tau_l$ and $\tau_h$, to generate pseudo-labels, where $\tau_h$ is usually set to a very high value to ensure that only reliable foreground pseudo-labels can participate in the supervision. In contrast, during the training of dual student framework, the CAMs tend to be more reliable gradually. Based on this intuition, to fully utilize more foreground pseudo-labels for sufficient training, we adjust the background threshold $\tau_h$ with the cosine descent strategy in every iteration: 
\begin{equation}
\tau_h(\mathrm{t})= \tau_h(0)-\frac{1}{2} \left(\tau_h(0)-\tau_h(\mathrm{T})\right)(1-\cos(\frac{\mathrm{t}\pi}{\mathrm{T}})),
\end{equation}
where $\mathrm{t}$ is the current number of iteration and $\mathrm{T}$ is the total number of training iterations.

\vspace{2pt}

\noindent\textbf{Adaptive Noise Filtering.} To further reduce the noise in the produced pseudo-labels that impacts the segmentation generalizability and reinforces the CAM confirmation bias, we develop an adaptive noise filtering strategy to implement \textit{trust-worthy} progressive learning. Previous studies suggest that deep networks tend to fit clean labels faster than noisy ones \cite{arpit2017closer,han2018co,ren2018learning}. This implies that the samples with smaller losses are more likely to be considered as the clean ones before the model overfits the noisy labels. A simple idea is to use a predefined threshold to divide the clean and noisy pseudo-labels based on their training losses. However, it fails to consider that the model's loss distribution is different across various samples, even those within the same class.

\vspace{1pt}

\begin{figure}[!tp]
    \centering
    \includegraphics[width=0.43\textwidth]{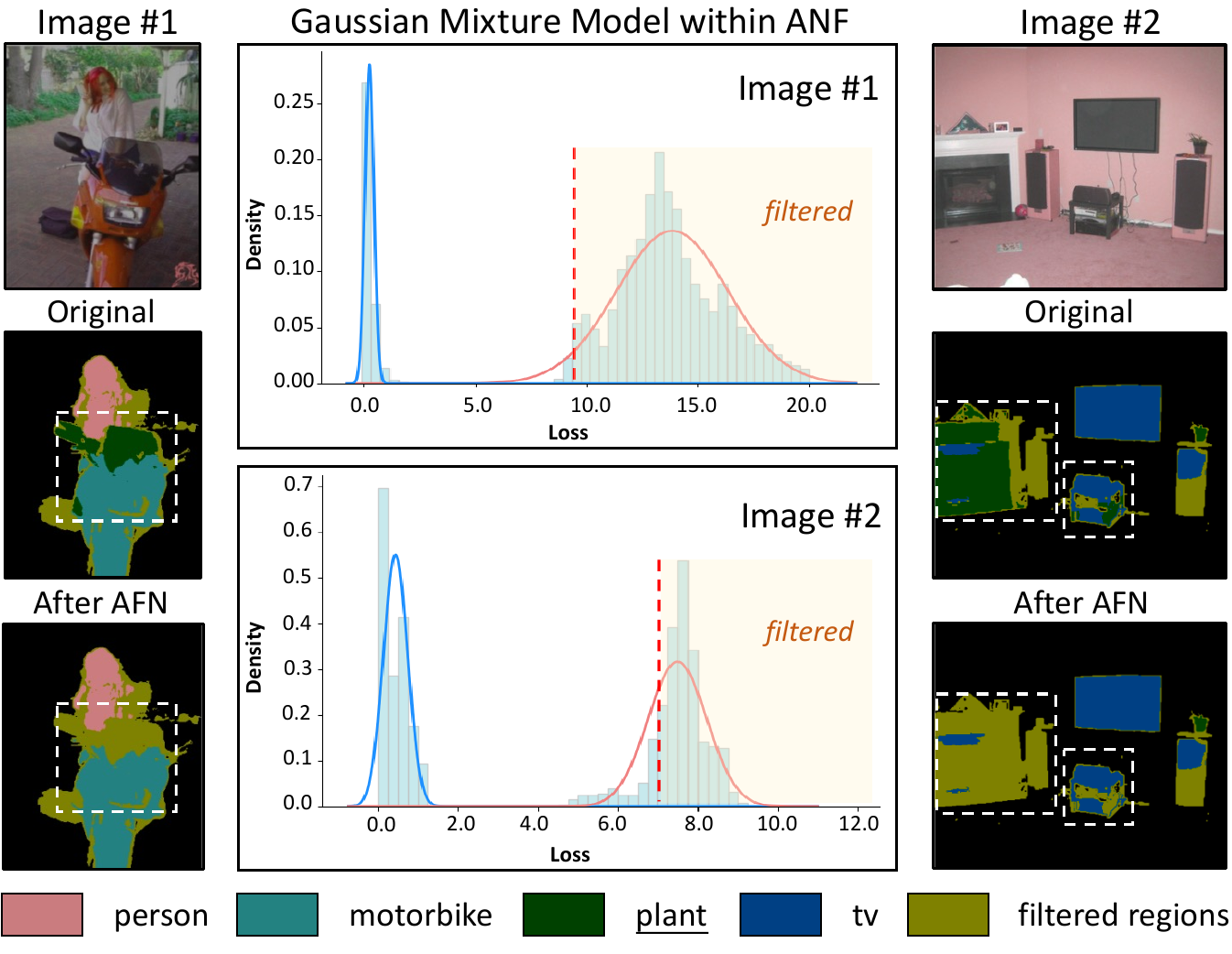}
    \vspace{-2mm}
    \caption{\textbf{The loss distribution of images with noisy pseudo-labels.} The model produces incorrect pseudo-labels of \texttt{plant}. Two peaks appear in the loss distribution on the two pseudo-labels, and the red peak with anomalous losses is mainly caused by noises. The distribution of normal losses is rescaled for visualization.}
    \vspace{-4mm}
    \label{fig_gmm}
\end{figure}

To this end, we develop an Adaptive Noise Filtering strategy to distinguish noisy and clean pseudo-labels via the loss distribution, as depicted in Figure \ref{fig_gmm}. Specifically, for the input image $\mathbf{X}$ with its segmentation map $\mathbf{P}$ and CAM pseudo-label $\mathbf{Y}$, we hypothesize the loss of each pixel $x \in \mathbf{X}$, defined as $l^x= \texttt{CE}\left(\mathbf{P}\left(x\right), \mathbf{Y}\left(x\right)\right)$, is sampled from a Gaussian mixture model (GMM) $\mathcal{P}(x)$ on all pixels with two components, \ie, clean $c$ and noisy $n$:
\begin{equation}
\mathcal{P}(l^x)=w_c \, \mathcal{N}(l^x | \mu_c,\left(\sigma_c\right)^2)+w_n \, \mathcal{N}(l^x | \mu_n,\left(\sigma_n\right)^2),
\end{equation}
where $\mathcal{N}(\mu, \sigma^2)$ represents one Gaussian distribution, $w_c, \mu_c, \sigma_c$ and $w_n, \mu_n, \sigma_n$ correspond to the weight, mean, and variance of two components. Thereinto, the component with high loss values corresponds to the noise component. Through Expectation Maximization algorithm \cite{li2020dividemix}, we can infer the noise probability $\varrho_n(l^x)$, which is equivalent to the posterior probability of $\mathcal{P}(noise \mid l^x, \mu_n, (\sigma_n)^2)$. If $\varrho_n(l^x) > \gamma$, the corresponding pixel will be classified as noise. Note that not all pseudo-labels $\mathbf{Y}$ are composed of noise, and thus the loss distributions may not have two clear Gaussian distributions. Therefore, we additionally measure the distance between $\mu_c$ and $\mu_n$. If $(\mu_n - \mu_c) \leq \eta$, all the pixel pseudo-labels will be regarded as clean ones. Finally, the set of noisy pixel pseudo-labels are determined as
\begin{equation}
\mathcal{X}_n=\left\{x \mid \varrho_n(l^x)>\gamma, \mu_c-\mu_n>\eta\right\},
\end{equation}
and they are excluded in the segmentation supervision. In DuPL, each sub-net's pseudo-labels are conducted adaptive noise filtering strategy independently.

\vspace{2pt}

\noindent\textbf{Every Pixel Matters.} In one-stage WSSS, discarding unreliable pseudo-labels that probably contain noises is a common practice to ensure the quality of the segmentation or other auxiliary supervision \cite{ru2022learning, xu2023self,ru2023token}. Although we gradually introduce more pixels to the segmentation training, there are still many unreliable pseudo-labels being discarded due to the semantic ambiguity of CAM. Typically, throughout the training phase, unreliable regions often exist in non-discriminative regions, boundaries, and background regions. Such an operation may cause the segmentation head to lack sufficient supervision in these regions.

To address this limitation, we treat the regions with unreliable pseudo-labels as \textit{unlabeled} samples. Despite no clear pseudo-labels to supervise the segmentation in these regions, we can regularize the segmentation head to output consistent predictions when fed perturbed versions of the same image. The consistency regularization implicitly imposes the model to comply with the smoothness assumption \cite{bachman2014learning,laine2016temporal}, which provides additional supervision for these regions. Specifically, we first apply strong augmentation $\phi$ to perturb the input image $\phi(\mathbf{X}) \rightarrow \widetilde{\mathbf{X}}$, and then send it to the sub-nets to get the segmentation prediction $\widetilde{\mathbf{P}}_i$ from $\psi^s_i$. Using the pseudo-label $\phi'(\mathbf{Y}_i)$ taking the same affine transformation in $\phi$ as the supervision, the consistency regularization of the $i$-th sub-net is formulated as:
\begin{equation}
\mathcal{L}_{reg\_i}=\frac{1}{\mid\mathcal{M}_i\mid} \sum_{x \in \mathbf{X}} \texttt{CE}\left[\widetilde{\mathbf{P}}_i(\phi(x)), \phi'(\mathbf{Y}_i(x))\right] \cdot\mathcal{M}_i,
\end{equation}
where $\mathcal{M}_i$ is the mask indicating the filtered pixels with unreliable pseudo-labels of the $i$-th sub-net. The filtered pixel is masked as $1$, and otherwise it is $0$. The total regularization loss of our dual student framework is $\mathcal{L}_{reg}=\mathcal{L}_{reg\_1} + \mathcal{L}_{reg\_2}$. This loss is computed for each image, with the total loss being the average across all images.

\subsection{Training objective of DuPL}
\label{sect_train_object}
As illustrated in Figure \ref{fig_overview}, DuPL consists four training objectives, that are, the classification loss $\mathcal{L}_{cls}$, the discrepancy loss $\mathcal{L}_{dis}$, the segmentation loss $\mathcal{L}_{seg}$, and consistency regularization loss $\mathcal{L}_{reg}$. Following the common practice in WSSS, we use the multi-label soft margin loss for classification. The total optimization objective of DuPL is the linear combination of the above loss terms:
\begin{equation}
\mathcal{L} = \mathcal{L}_{cls} + \lambda_1\mathcal{L}_{dis} + \lambda_2\mathcal{L}_{seg} + \lambda_3\mathcal{L}_{reg},
\end{equation}
where $\lambda_i$ is the weight to rescale the loss terms.

\begin{table}[tbp]
  \centering
  \renewcommand\arraystretch{1.1}
  \setlength{\tabcolsep}{2.3mm}
  \small
  {
    \begin{tabular}{l|c|ccc}
      \toprule[1.2pt]
      {\textbf{Method}}                                           & \textbf{Backbone}        & \texttt{train} & \texttt{val}  \\\midrule
      \multicolumn{4}{l}{\textit{\textbf{Multi-stage WSSS Methods}}} \\
      PPC \cite{du2022weakly} \begin{tiny}CVPR'2022\end{tiny} + PSA \cite{ahn2018learning}& WR38 & \,\,\,73.3 & -- \\
      ACR  \cite{kweon2023weakly} \begin{tiny}CVPR'2023\end{tiny} + IRN \cite{ahn2019weakly}& WR38 & 72.3 & -- \\
      \midrule
      \multicolumn{4}{l}{\textit{\textbf{One-stage WSSS Methods}}} \\
1Stage \cite{araslanov2020single}  \tiny CVPR'2020 & WR38            & 66.9           & 65.3          \\
ViT-PCM \cite{rossetti2022max} \tiny ECCV'2022          & \,\,\,ViT-B$^\dagger$ & 67.7           & 66.0          \\
      AFA \cite{ru2022learning} \tiny CVPR'2022          & MiT-B1          & 68.7           & 66.5          \\
      {ToCo \cite{ru2023token}} \tiny CVPR'2023                                      & ViT-B           & 72.2           & 70.5          \\
      \rowcolor[HTML]{F7E0D5}
      \textbf{DuPL} & ViT-B & 75.1 & 73.5 \\
      \rowcolor[HTML]{F7E0D5}
      \textbf{DuPL$^\dagger$} & \,\,\,ViT-B$^\dagger$ & \textbf{76.0} & \textbf{74.1} \\
      \bottomrule[1.2pt] 
    \end{tabular}
    \vspace{-1mm}
    \caption{\textbf{Evaluation of CAM pseudo labels.} The results are evaluated on the VOC \texttt{train} and \texttt{val} set and reported in mIoU (\%). $\dagger$ denotes using ImageNet-21k pretrained parameters.}
    \label{tab_pseudo_label}%
  }
  \vspace{-4mm}
\end{table}

\section{Experiments}
\subsection{Experimental Settings}

\noindent\textbf{Datasets.} We evaluate the proposed DuPL on the two standard WSSS datasets, \ie, PASCAL VOC 2012 and MS COCO 2014 datasets. For the VOC 2012 dataset, it is extended with the SBD dataset \cite{hariharan2011semantic} following common practice. The train, val, and test set are composed of 10582, 1449, and 1456 images, respectively. The test performance of DuPL is evaluated on the official evolution server. For the COCO 2014 dataset, its train and val set involve 82k and 40k images, respectively. The mean Intersection-over-Union (mIoU) is reported for performance evaluation.

\vspace{2pt}

\noindent\textbf{Network Architectures of DuPL.} We use the ViT-B \cite{dosovitskiy2020image} with a lightweight classifier and a segmentation head, and the patch token contrast loss \cite{ru2023token} as our baseline network. The classifier is a fully connected layer. The segmentation head consists of two $3 \times 3$ convolutional layers (with a dilation rate of $5$) and one $1 \times 1$ prediction layer. The patch token contrast loss is applied to alleviate the over-smoothness issue of CAM in ViT-like architectures. DuPL is composed of two subnets with the baseline settings, where the backbones are initialized with ImageNet pretrained weights.

\vspace{2pt}

\noindent\textbf{Implement Details.} We adopt the AdamW optimizer with an initial learning rate set to $6e^{-5}$ and a weight decay factor $0.01$. The input images are augmented using the strategy in \cite{ru2023token}, and cropped to $448 \times 448$. For the strong perturbations, we adopt Random Augmentation strategy \cite{cubuk2020randaugment} on color and apply additional scaling and horizontal flipping. In the inference stage, following the common practice in WSSS, we use multi-scale testing and dense CRF processing.

For experiments on the VOC 2012 dataset, the batch size is set as $4$. The total iteration is set as 20k with 2k iterations warmed up for the classifiers and 6k iterations warmed up for the segmentation heads before conducting Adaptive Noise Filtering. The background thresholds $(\tau_l, \tau_h(0), \tau_h(\mathrm{T}))$ are set as ($0.25, 0.7, 0.55$). The thresholds ($\gamma,\eta$) of Adaptive Noise Filtering are set as ($0.9, 1.0$). The weight factors ($\lambda_1,\lambda_2,\lambda_3$) of the loss terms in Section \ref{sect_train_object} are set as ($0.1, 0.1, 0.05$). For the COCO dataset, the batch size is set as $8$. The network is trained for 80k iterations with 5k iterations warmed up for the classifier, and 20k iterations warmed up for the segmentation head. The other settings are remained the same.

\begin{figure}[!tp]
    \centering
    \includegraphics[width=0.45\textwidth]{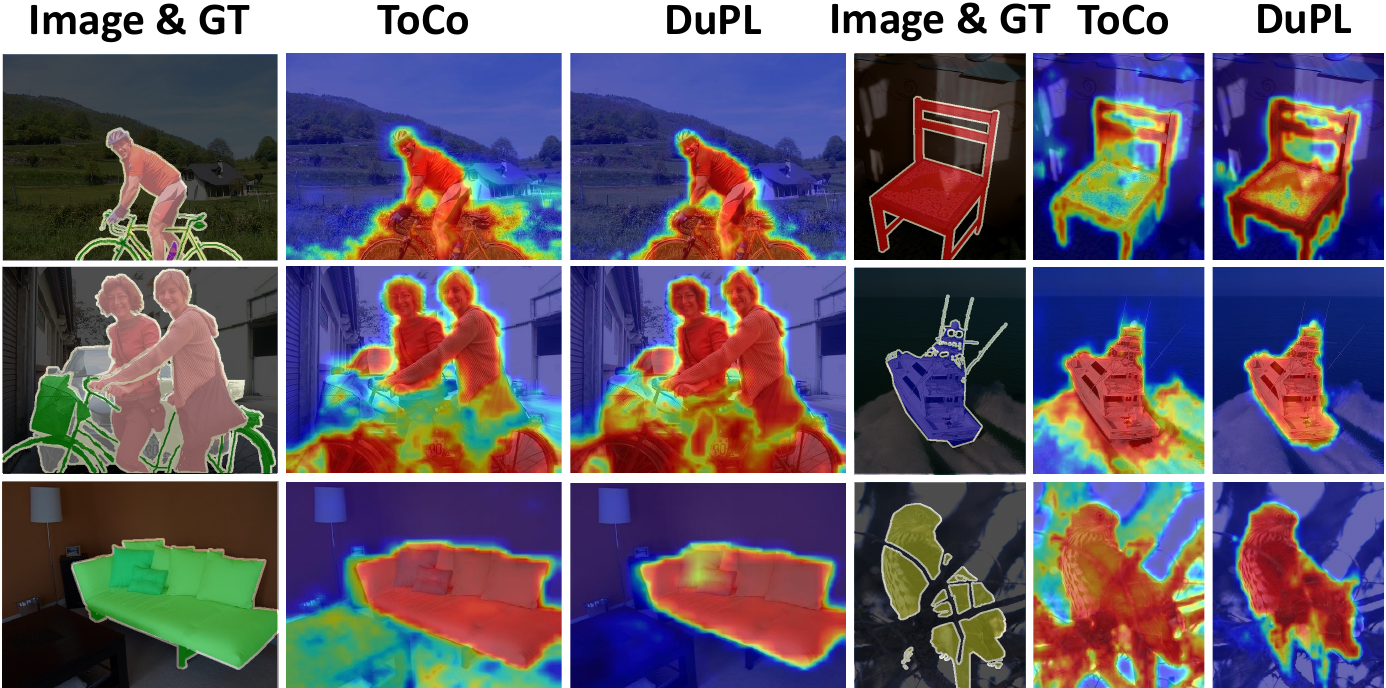}
    \vspace{-3pt}
    \caption{\textbf{Visual comparison of CAMs.} We compare the state-of-the-art one-stage approach, ToCo \cite{ru2023token}, with our proposed DuPL. DuPL not only suppresses over-activations but also achieves more complete object activation coverage.} 
    \vspace{-5mm}
    \label{fig_cam}
\end{figure}

\subsection{Experimental Results}
\noindent\textbf{CAM and Pseudo-labels.} We begin by visualizing the CAM of DuPL in Figure \ref{fig_cam}. We can find that, using the same ViT-B backbone with ImageNet-1k pretrained weights, our method can generate more complete and accurate CAMs when compared to current state-of-the-art one-stage work, \ie, ToCo \cite{ru2023token}. Then, we evaluate the CAM pseudo-labels on the \texttt{train} and \texttt{val} set of the VOC dataset and compare them with recent state-of-the-art WSSS methods. In one-stage methods, the pseudo-labels are directly generated using CAMs, while those of multi-stage methods are produced by the initial seed generation and refinement processes. The results are presented in Table \ref{tab_pseudo_label}. As can be seen, DuPL significantly outperforms the recent one-stage competitors and even surpasses the multi-stage methods. Compared with other methods with ViT-B baseline, our methods can produce higher quality pseudo-labels than the competitors with both the ImageNet-1k and ImageNet-21k pretrained weights. Using ViT-B with ImageNet-21k pretrained weights, we boost the pseudo-label performance to 76.0\% (\textbf{+3.8\%}) and 74.1\% (\textbf{+3.6\%}) on the \texttt{train} and \texttt{val} set, respectively. 

\vspace{2pt}

\begin{table}[tbp]
  \small
  \centering
  \renewcommand\arraystretch{1.05}
  \setlength{\tabcolsep}{0.8mm}
  \begin{tabular}{l|c|c|cc|c}
    \toprule[1.2pt]
& \multirow{2}{*}{\textbf{Sup.}}   & \multirow{2}{*}{\textbf{Net.}  } & \multicolumn{2}{c|}{\textbf{VOC}} & {\textbf{COCO}}                                                                                           \\ \cmidrule{4-6}
&                           &                           & \texttt{val}                      & \texttt{test}                                                                             & \texttt{val}  \\ \midrule
    \multicolumn{4}{l}{\cellcolor[HTML]{ffffff}\textbf{\textit{Multi-stage WSSS Methods}}.}                                                                                                                                                                   \\
    EPS \cite{lee2021railroad}  \tiny CVPR'2021       & $\mathcal{I}+\mathcal{S}$ & DL-V2                     & 71.0                              & 71.8                                                                                      & --            \\
    L2G \cite{jiang2022l2g} \tiny CVPR'2022           & $\mathcal{I}+\mathcal{S}$ & DL-V2                     & 72.1                              & 71.7                                                                                      & 44.2          \\
    PPC  \cite{du2022weakly}   \tiny CVPR'2022  & $\mathcal{I}+\mathcal{S}$ & DL-V2                     & 72.6                              & 73.6                                                                                      & --            \\
        Lin \etal \cite{lin2023clip} \tiny CVPR'2023 & $\mathcal{I}+\mathcal{T}$  & DL-V2 & 71.1 & 71.4 & 45.4 \\ 
    ReCAM \cite{chen2022class} \tiny CVPR'2022        & $\mathcal{I}$             & DL-V2                     & 68.4                              & 68.2                                                                                      & 45.0          \\
    W-OoD \cite{lee2022weakly} \tiny CVPR'2022        & $\mathcal{I}$             & WR-38                      & 70.7                              & 70.1                                                                                      & --            \\
       ESOL \cite{li2022expansion} \tiny NeurIPS'2022    & $\mathcal{I}$             & DL-V2                     & 69.9                              & 69.3                                                                                      & 42.6 \\ 
    MCTformer \cite{xu2022multi} \tiny CVPR'2022      & $\mathcal{I}$             & WR-38                      & 71.9                              & 71.6                                                                                      & 42.0          \\        
    OCR \cite{cheng2023out} \tiny CVPR'2023 & $\mathcal{I}$ & WR-38 & 72.7 & 72.0 & 42.5 \\
    ACR \cite{kweon2023weakly} \tiny CVPR'2023     & $\mathcal{I}$ & DL-V2                     & 71.9                              & 71.9                                                                                      & 45.3   \\
    \midrule
    \multicolumn{4}{l}{\cellcolor[HTML]{ffffff}\textbf{\textit{One-stage WSSS Methods}}.}                                                                                                                                                                  \\
    RRM \cite{zhang2020reliability} \tiny AAAI'2020   & $\mathcal{I}$             & WR-38                      & 62.6                              & 62.9                                                                                      & --            \\
    1Stage \cite{araslanov2020single} \tiny CVPR'2020 & $\mathcal{I}$             & WR-38                      & 62.7                              & 64.3                                                                                      & --            \\
    AFA \cite{ru2022learning} \tiny CVPR'2022           & $\mathcal{I}$             & MiT-B1                    & 66.0                              & 66.3                                                                                      & 38.9          \\
    SLRNet \cite{pan2022learning} \tiny IJCV'2022     & $\mathcal{I}$             & WR-38                      & 67.2                              & 67.6                                                                                      & 35.0          \\
    TSCD \cite{xu2023self} \tiny AAAI'2023 & $\mathcal{I}$ & MiT-B1 & 67.3 & 67.5 & 40.1 \\
    ToCo \cite{ru2023token} \tiny CVPR'2023                                   & $\mathcal{I}$             & ViT-B                     & 69.8                              & 70.5          & 41.3          \\
    \rowcolor[HTML]{F7E0D5}
    \textbf{DuPL}                                     & $\mathcal{I}$             & ViT-B                     & 72.2                              & \,\,71.6\tablefootnote{\url{http://host.robots.ox.ac.uk:8080/anonymous/103D8M.html}}          & 43.5          \\
    \rowcolor[HTML]{F7E0D5}
    \textbf{DuPL$^\dagger$}                                     & $\mathcal{I}$             & \,\,\,ViT-B$^\dagger$                     & \textbf{73.3}                              & \,\,\textbf{72.8}\tablefootnote{\url{http://host.robots.ox.ac.uk:8080/anonymous/R7RLMS.html}}          & \textbf{44.6}          \\
 \bottomrule[1.2pt]
  \end{tabular}
  \vspace{-1mm}
  \caption{\textbf{Semantic Segmentation Results}. “Sup.” denotes the supervision type. $\mathcal{I}$: Image-level labels; $\mathcal{S}$: Saliency maps. $\mathcal{T}$: text-driven supervision from CLIP \cite{radford2021learning}. “Net.” denotes the backbone in one-stage methods and the segmentation network in multi-stage methods. $\dagger$ denotes using ImageNet-21k pretrained weights.}
  \label{tab_seg}
  \vspace{-16pt}
\end{table}

\begin{figure*}[!tp]
    \centering
    \includegraphics[width=0.8\textwidth]{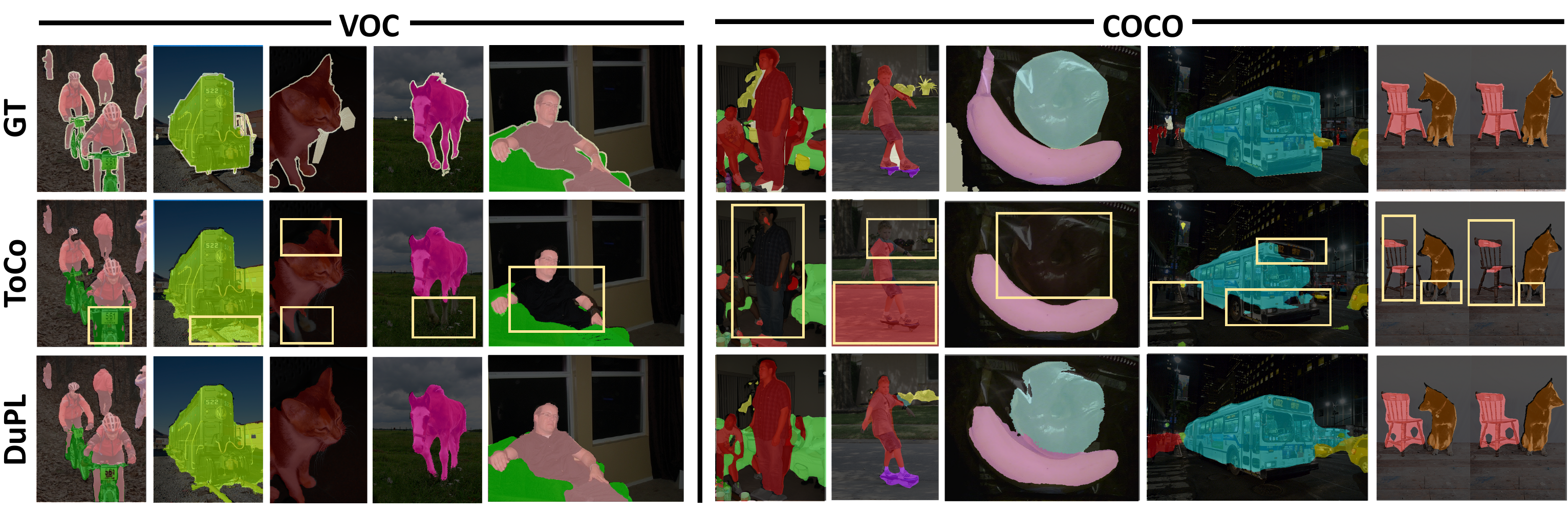}
    \vspace{-2pt}
    \caption{\textbf{Visualization of segmentation results on PSCAL VOC 2012 and MS COCO datasets}. We compare the results of DuPL with those of ToCo \cite{ru2023token}. Both of them use ViT-B with ImageNet-1k as the backbone for fair comparison.} 
    \vspace{-3mm}
    \label{fig_seg}
\end{figure*}

\vspace{2pt}

\noindent\textbf{Final Segmentation Results.} Table \ref{tab_seg} reports the final segmentation performance of DuPL. To show the superiority of the proposed method, we compare our performance with both one-stage and multi-stage prior arts. Notably, the proposed DuPL achieves 73.3\% (\textbf{+3.5\%}), 72.8\% (\textbf{+2.3\%}) and 44.6\% (\textbf{+3.3\%}) mIoU on the VOC \texttt{val}, \texttt{test} and COCO \texttt{val} set, respectively, which significantly surpasses recent one-stage methods. The performance of DuPL strongly supports that \textit{fully exploiting the trustworthy pseudo-labels is very important to single-stage methods}. Also, DuPL proves that using the one-stage pipeline is \textit{strong enough} to achieve competitive WSSS performance with multi-stage approaches. Along with the quantitative comparison results, we visualize and compare the segmentation masks of DuPL, ToCo \cite{ru2023token}, and ground-truths in Figure \ref{fig_seg}. We can see that DuPL predicts more accurate objects in challenging scenes, which are close to their ground truths. 

\vspace{2pt}

\noindent\textbf{Fully-Supervised Counterparts.} As presented in Table \ref{tab:full_compare}, the one-stage competitors adopt various backbones, \eg., Wide ResNet38 (WR-38), MixFormer-Base1 (MiT-B1), and ViT-Base (ViT-B). To eliminate the impact of backbone on segmentation results for fair comparison, we compared the performance gap between the methods and their fully supervised counterpart. Notably, when using the ImageNet-1k pre-trained weight, DuPL achieves 72.2\% mIoU and \textbf{90.1\%} of its upper bound performance, significantly ahead of recent one-stage one-stage methods (\textbf{+3.4\%}). 

\subsection{Ablation studies and Analysis}
\noindent\textbf{Effectiveness of Components.} The proposed DuPL consists of a dual student (DS) architecture and trust-worthy progressive learning. Within the progressive learning, we have dynamic threshold adjustment (DTA) and Adaptive Noise Filtering (ANF). In addition to the basic classification and segmentation loss, DuPL also incorporates two training losses, \ie, discrepancy loss $\mathcal{L}_{dis}$ and consistency regularization loss $\mathcal{L}_{reg}$. We now investigate the contributions of each module and loss in DuPL. 

The experiment results are presented in Table \ref{tab_ablate}. We can observe that employing solely dual student architecture brings a slight improvement of nearly 2\% mIoU for CAM pseudo-labels, resulting in 62.3\% mIoU of segmentation performance. In this setting, the CAM diversity arises merely from the randomly initialized segmentation heads, thus the CAMs from the two sub-nets are still highly identical, leaving a huge space for improvement. When incorporating $\mathcal{L}_{dis}$, the performance of CAM pseudo-label is improved to 67.3\% mIoU, indicating that it can further benefit the effectiveness of dual student architecture. As CAM becomes increasingly reliable, DTA progressively introduces more pixels into the segmentation supervision and improves the segmentation performance by 2.6\%. The ANF suppresses noise pseudo-labels and improves segmentation performance by 1.5\%. It’s noted that high-quality supervision of segmentation benefits the CAM quality, and DTA with ANF significantly improves the pseudo labels by 4.3\%. With the motivation of “every pixel matters”, $\mathcal{L}_{dis}$ ultimately boosts the segmentation performance to 69.9\% mIoU, leading to the state-of-the-art.

\begin{table}[t]
\small
\centering
\renewcommand\arraystretch{1.05}
\setlength{\tabcolsep}{2.2mm}{
\begin{tabular}{l|c|cc|c}
\toprule[1.2pt]
\textbf{Method} & \textbf{BB.} & \texttt{val} ($\mathcal{F}$) & \texttt{val} ($\mathcal{I}$) & \textit{ratio (\%)} \\ \midrule
1Stage \cite{araslanov2020single}          & WR38         & 80.8         & 62.7         & 77.6         \\
SLRNet \cite{pan2022learning}          & WR38         & 80.8         & 67.2         & 83.2         \\
AFA \cite{ru2022learning}             & MiT-B1       & 78.7         & 66.0         & 83.9         \\ 
ToCo \cite{ru2023token}            & ViT-B       & 80.5         & 69.8         & 86.7         \\
\midrule
\rowcolor[HTML]{F7E0D5} 
\textbf{DuPL}     & \textbf{ViT-B}       & \textbf{80.5}         & \textbf{72.2}         & \textbf{90.1}         \\
\rowcolor[HTML]{F7E0D5}
\textbf{DuPL$^\dagger$}     & \textbf{\,\,\,ViT-B$^\dagger$}       & \textbf{82.3}         & \textbf{73.3}         & \textbf{89.1}         \\ 
\bottomrule[1.2pt]
\end{tabular}}
\vspace{-1.5mm}
\caption{\textbf{The performance comparison with fully supervised counterparts on the VOC dataset.} The pixel pseudo labels are used to supervise the segment head. $\mathcal{F}$: fully-supervised supervision. $\mathcal{I}$: image-level supervision (WSSS). \textit{ratio} = \texttt{val} ($\mathcal{I}$) / \texttt{val} ($\mathcal{F}$). $\dagger$ denotes using ImageNet-21k pretrained weights.}
\label{tab:full_compare}
\vspace{-2mm}
\end{table}

\begin{table}[htb]
\small
\renewcommand\arraystretch{1.05}
\setlength{\tabcolsep}{1.94mm}
\begin{tabular}{c|cc|ccc|c|c}
\toprule[1.2pt]
\textbf{Baseline} & \textbf{DS} & $\mathbf{\mathcal{L}_{dis}}$ & \textbf{DTA} & \textbf{ANF} & $\mathbf{\mathcal{L}_{reg}}$ & $\mathbf{M}$ & \textbf{Seg.} \\ \midrule
\checkmark        &              &           &         &     &          & 63.2     & 62.3     \\ \midrule
\checkmark        & \checkmark            &           &         &     &          & 65.4     & 63.8     \\
\checkmark        & \checkmark            & \checkmark         &         &     &          & 67.3     & 64.1     \\ \midrule
\checkmark        & \checkmark            & \checkmark         & \checkmark       &     &          & 69.2     & 66.7     \\
\checkmark        & \checkmark            & \checkmark         & \checkmark       & \checkmark   &          & 71.6     & 68.2     \\
\rowcolor[HTML]{F7E0D5}
\checkmark        & \checkmark            & \checkmark         & \checkmark       & \checkmark   & \checkmark        & \textbf{73.5}     & \textbf{69.9}     \\ \bottomrule[1.2pt]
\end{tabular}
\vspace{-1mm}
  \caption{\textbf{Ablation Study.} “$\mathbf{M}$” denotes the CAM performance and “Seg.” denotes the segmentation performance. CRF post-processing is not conducted in the ablation study.}
  \label{tab_ablate}
  \vspace{-5mm}
\end{table}

\vspace{4pt}

\begin{figure}
\centering
    \begin{subfigure}{0.475\textwidth}
        \centering
        \includegraphics[width=0.94\textwidth]{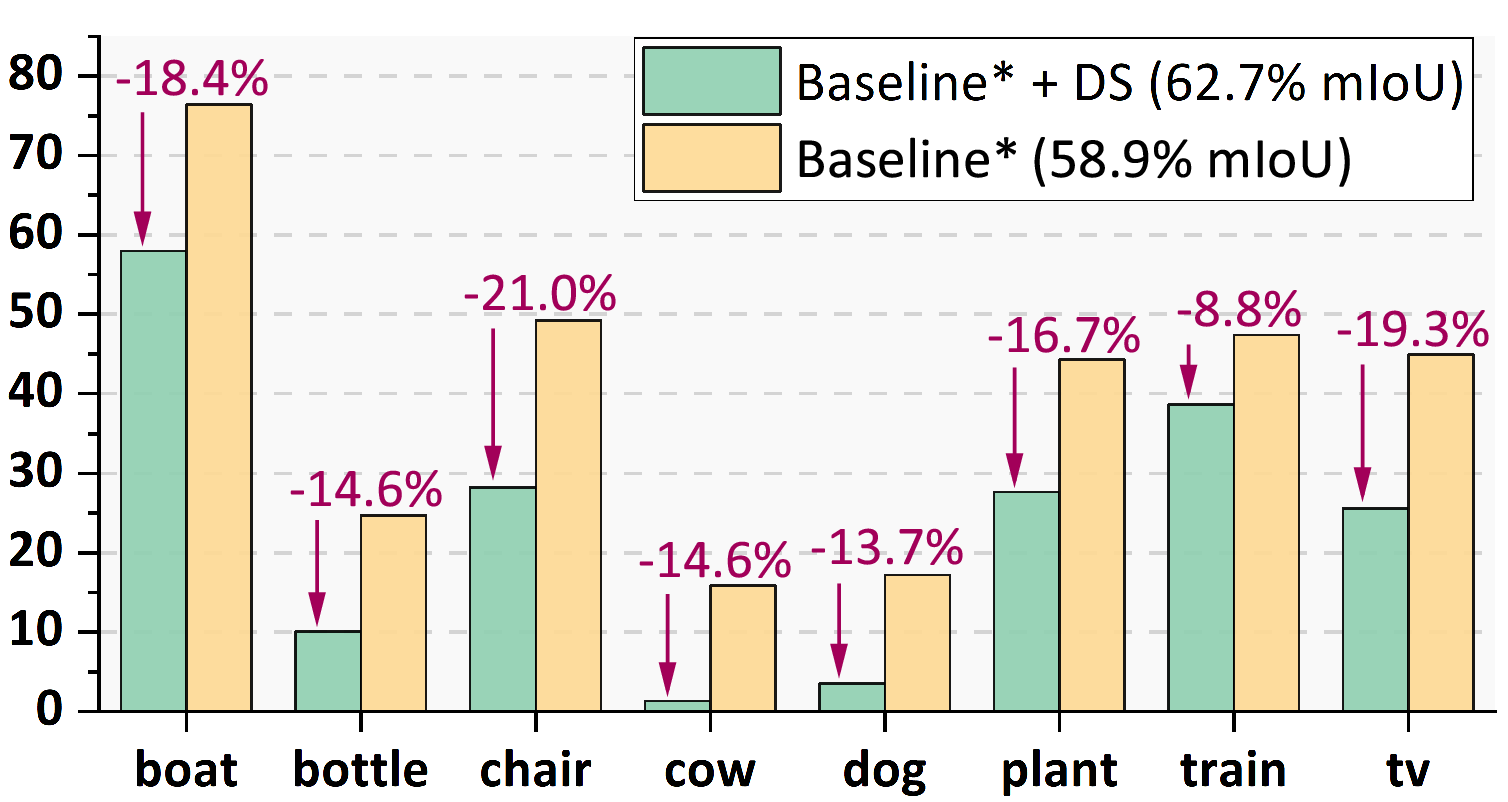}
        \caption{Comparison of the baseline and the baseline with dual student.}
        \label{pic_oa_1}
        \vspace{4pt}
    \end{subfigure}
    \begin{subfigure}{0.475\textwidth}
        \centering
        \includegraphics[width=0.94\textwidth]{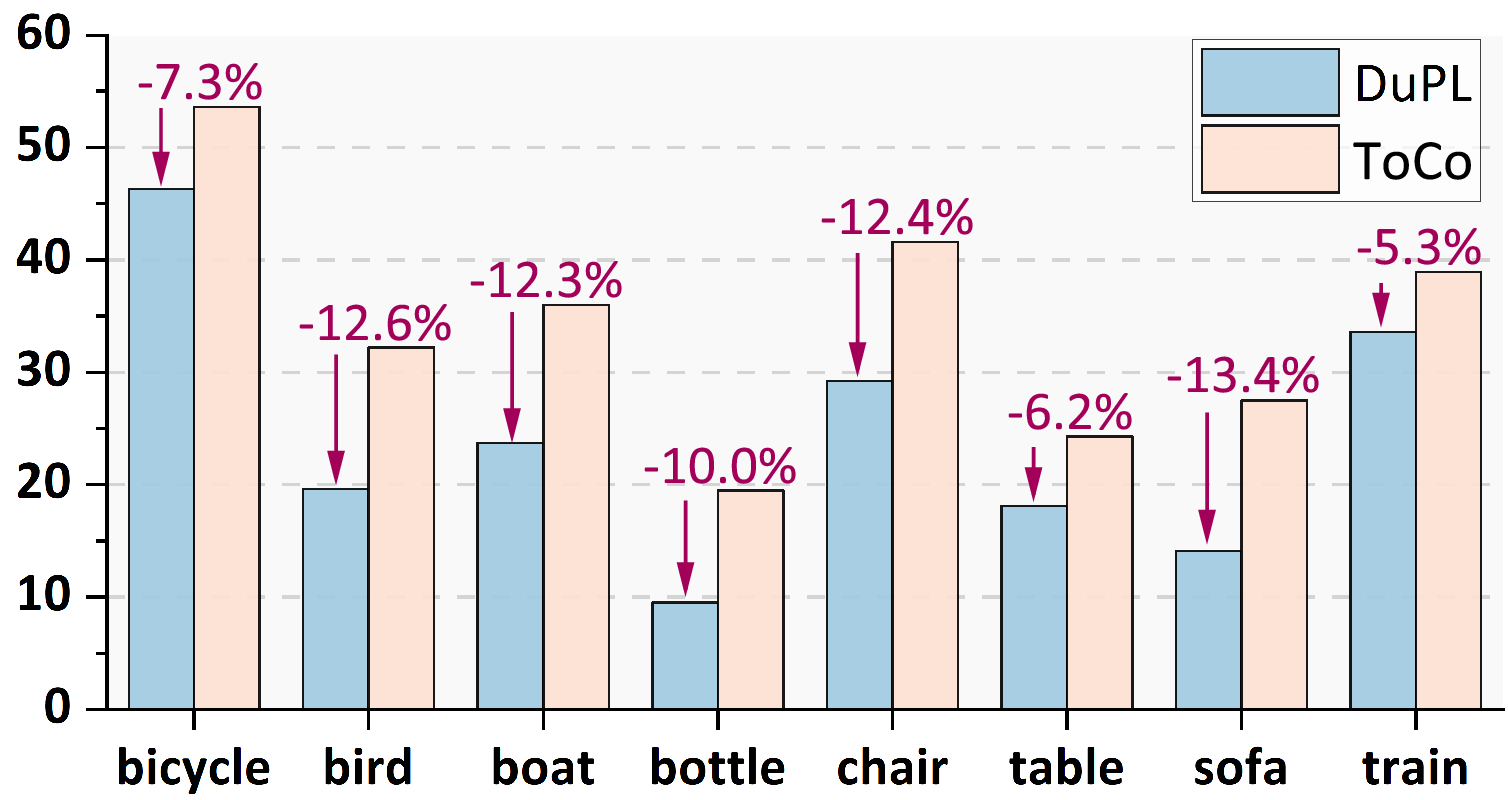}
        \caption{Comparison of ToCo \cite{ru2023token} and the proposed DuPL.}
        \label{pic_oa_2}
    \end{subfigure}
    \vspace{-16pt}
    \caption{\textbf{Effectiveness evaluation of our proposed method.} The OA rate (\%) are evaluated on the VOC \texttt{val} set. “*” denotes the baseline is trained under a low background threshold ($\tau_h=0.5$) to aggregate the CAM conformation bias. The per-class results can be viewed in \textit{Supplementary Material.}}
    \label{pic_oa}
    \vspace{-5mm}
\end{figure}

\noindent\textbf{Analysis of Dual Student.} DuPL adopts the mutual supervision of two student sub-nets to alleviate the confirmation bias introduced by their own incorrect pseudo-labels. The confirmation bias issue can be reflected by the over-activation (OA) rate. A higher OA rate means the model activates more incorrect pixels for the target classes, causing a more severe CAM confirmation bias. Here, we count the number of the false positive (FP) and true positive (TP) pixel pseudo-labels for each class, and calculate the OA rate (\ie, \texttt{FP} / (\texttt{TP} + \texttt{FP})). We first compare the baseline and the ablated variant with Dual Student (\ie, baseline + DS + $\mathcal{L}_{dis}$) under a low background threshold setting ($\tau_h=0.5$). From Figure \ref{pic_oa_1}, we can see due to the confirmation bias, the baseline over-activates lots of incorrect regions, resulting in subpar segmentation outcomes (only 58.9\% mIoU). With dual student, the ablated version significantly reduces the OA rate by over 15\% in many classes, and even reduces the OA rate to below 5\% for some categories (such as \texttt{cow} and \texttt{dog}). Further, we evaluate the OA rate of ToCo \cite{ru2023token} and DuPL. From Figure \ref{pic_oa_2}, we can see that ToCo also suffers from the confirmation bias problem, with OA rate exceeding 30\% in several categories. In contrast, the proposed DuPL significantly overcomes this problem in these severely over-activated classes, which reflects the effectiveness of our architecture.

\vspace{2pt}

\noindent\textbf{Dynamic Threshold Adjustment.} In DuPL, $\tau_h(\mathrm{t})$ is a dynamic background threshold that progressively decreases to $\tau_h(\mathrm{T})$ with training, aiming at involving more pseudo-labels into the segmentation supervision. Table \ref{tab_ablate_dta} shows the impact of different $\tau_h(\mathrm{T})$ on the CAM and segmentation performance. We observe that when $\tau_h(\mathrm{T})$ ranges from 0.65 to 0.55, the model's performance exhibits steady improvement. However, when $\tau_h(\mathrm{T})$ is smaller than $0.55$, the excessive introduction of noises becomes challenging to suppress, thus yielding a negative impact on the model performance. Nevertheless, the model continues to improve in comparison to the case with a relatively higher $\tau_h(\mathrm{T})$.

\vspace{2pt}

\noindent\textbf{Warm-up Stage for The Segmentation Head.} Motivated by the Early-learning nature of deep networks, ANF uses the feedback from the segmentation head to filter the noise pseudo-labels. This requires the segmentation head to fit the CAM pseudo-labels properly. Incorporating ANF too early may risk filtering out correct pseudo-labels due to under-fitting, while introducing ANF too late may lead to the model having already memorized noisy pseudo-labels, making it challenging to discriminate them. In Table \ref{tab_ablate_warm}, we report the impact on the warm-up stage for the segmentation head. We show that warming up the segmentation head using 8000 iterations can achieve the best performance.

\begin{table}[t]
\small
\centering
\begin{subfigure}{0.235\textwidth}
\renewcommand{\arraystretch}{1.03}
\setlength{\tabcolsep}{3.1mm}
\centering
\begin{subtable}[t]{1.0\textwidth}
\begin{tabular}{c|c|c}
\toprule[1.2pt]
$\mathbf{\tau_h(T)}$                 & $\mathbf{M}$                         & \textbf{Seg.}                           \\ \midrule
0.65                                 & 69.4                                 & 68.1                                 \\
0.60                                  & 71.8                                 & 70.9                                  \\
\rowcolor[HTML]{F7E0D5} 
{\color[HTML]{333333} \textbf{0.55}} & {\color[HTML]{333333} \textbf{73.5}} & {\color[HTML]{333333} \textbf{72.2}} \\
0.50                                  & 72.3                                 & 71.5                              
\\ \bottomrule[1.2pt]
\end{tabular}
\vspace{2pt}
\caption{Background threshold $\tau_h$.}
\label{tab_ablate_dta}
\end{subtable}
\end{subfigure}
\begin{subfigure}{0.235\textwidth}
\renewcommand{\arraystretch}{1.03}
\setlength{\tabcolsep}{3.6mm}
\centering
\begin{subtable}[t]{1.0\textwidth}
\begin{tabular}{c|c|c}
\toprule[1.2pt]
\textbf{Iter}                         & $\mathbf{M}$                         & \textbf{Seg.}                            \\ \midrule
6000                                  & 72.4                                 & 70.9                                 \\
\rowcolor[HTML]{F7E0D5}
\textbf{8000}                         & \textbf{73.5}                        & \textbf{72.2}                        \\ 
10000                                 & 72.6                                 & 71.7                                 \\
12000                                 & 71.1                                 & 69.4                                 \\ \bottomrule[1.2pt]
\end{tabular}
\vspace{2pt}
\caption{Warm-up stage.}
\label{tab_ablate_warm}
\end{subtable}
\end{subfigure}
\vspace{-8pt}
\caption{\textbf{Impact of hyper-parameters.} The results are evaluated on the VOC \texttt{val} set. The default settings are marked in \colorbox[HTML]{F7E0D5}{color}.}
\vspace{-2pt}
\end{table}

\begin{table}[t]
\small
\centering
\renewcommand{\arraystretch}{1.05}
\setlength{\tabcolsep}{3mm}
\begin{tabular}{c|cccc}
\toprule[1.2pt]
\textbf{} & \textbf{None} & \textbf{Diff. Aug} & $\mathbf{\mathcal{L}_{dis}}$                                     & \textbf{Diff. Aug +} $\mathbf{\mathcal{L}_{dis}}$ \\ \midrule
\textbf{$\mathbf{M}$}    & 69.6 & 70.7      & \cellcolor[HTML]{F7E0D5}\textbf{73.5} & 70.9          \\
\textbf{Seg.}   & 68.9 & 69.8      & \cellcolor[HTML]{F7E0D5}\textbf{72.2} & 69.4          \\ \bottomrule[1.2pt]
\end{tabular}
\vspace{-3pt}
\caption{\textbf{Different discrepancy strategies in Dual student.} The results are evaluated on the VOC \texttt{val} set. “Diff. Aug” denotes that the input images of two-subnets are augmented differently, and the CAM pseudo-labels will be re-transformed to fit the inputs for the other sub-net.}
\label{tab_ablate_dis}
\vspace{-14pt}
\end{table}
 \vspace{4pt}
\noindent\textbf{Discrepancy strategy in Dual Student.} We apply the discrepancy constraint on the representation level to make each sub-nets generate more diverse CAMs. In Table \ref{tab_ablate_dis}, we compare the impact of different discrepancy strategies. It shows that only introducing $\mathcal{L}_{dis}$ on the representation level is more beneficial for two sub-nets to transfer the knowledge learned from one view to the other through CAM pseudo-labels, thus yielding favorable performance.

\vspace{-2pt}

\section{Conclusion}

This work aims to address the problem of CAM confirmation bias and fully utilize the CAM pseudo-labels for better WSSS. Specifically, we develop a dual student architecture with two sub-nets that mutually provide the pseudo-labels for the other, which is empirically proved to counter the CAM confirmation bias well. With better CAM activations during the training process, we gradually introduce more pixels into the supervision for sufficient segmentation training. We overcome the excessive noisy pseudo-labels brought by the above operation by proposing an adaptive noise filter strategy. Such a trustworthy progressive learning paradigm significantly boosts the WSSS performance. Motivated by the idea that “every pixel matters”, instead of discarding unreliable labels, we fully leverage them through consistency regularizations. The experiment results demonstrate that DuPL significantly outperforms other one-stage competitors and archives competitive performance with multi-stage solutions.

\vspace{8pt}

\noindent \textbf{Acknowledgements.} This work is supported in part by Shanghai science and technology committee under grant No. 22511106005. We appreciate the High Performance Computing Center of Shanghai University, and Shanghai Engineering Research Center of Intelligent Computing System for the computing resources and technical support.

\newpage

{
    \small
    \bibliographystyle{ieeenat_fullname}
    \bibliography{main}
}


\end{document}